\documentclass[11pt]{article}

\usepackage[final]{acl}

\usepackage{times}
\usepackage{latexsym}
\usepackage[T1]{fontenc}
\usepackage[utf8]{inputenc}
\usepackage{microtype}
\usepackage{inconsolata}
\usepackage{graphicx}

\newcommand{\crr}{\textrm{cr}}
\newcommand{\drr}{\textrm{rd}}

\usepackage{amsmath,amssymb,amsfonts}
\usepackage{algorithmic}
\usepackage{graphicx}
\usepackage{textcomp}
\usepackage{xcolor}
\usepackage{booktabs}
\usepackage{comment}
\usepackage{multirow}
\usepackage{makecell}
\usepackage{scrextend}
\usepackage{hyperref}

\title{Optimal Multi-Task Learning at Regularization Horizon \\ for Speech Translation Task}

\author{JungHo Jung \\
  University of Pennsylvania \\
  \texttt{j76jung@seas.upenn.edu} \\\And
  Junhyun Lee \\
  Samsung Research \\
  \texttt{junhyun8.lee@samsung.com} \\}

\begin{document}
\maketitle
\begin{abstract}
End-to-end speech-to-text translation typically suffers from the scarcity of paired speech-text data. One way to overcome this shortcoming is to utilize the bitext data from the Machine Translation (MT) task and perform Multi-Task Learning (MTL). In this paper, we formulate MTL from a regularization perspective and explore how sequences can be regularized within and across modalities. By thoroughly investigating the effect of consistency regularization (different modality) and R-drop (same modality), we show how they respectively contribute to the total regularization. We also demonstrate that the coefficient of MT loss serves as another source of regularization in the MTL setting. With these three sources of regularization, we introduce the optimal regularization contour in the high-dimensional space, called the regularization horizon. Experiments show that tuning the hyperparameters within the regularization horizon achieves near state-of-the-art performance on the MuST-C dataset. 
\end{abstract}

\begin{figure}[t]
  \centering
  \includegraphics[width=0.9\linewidth]{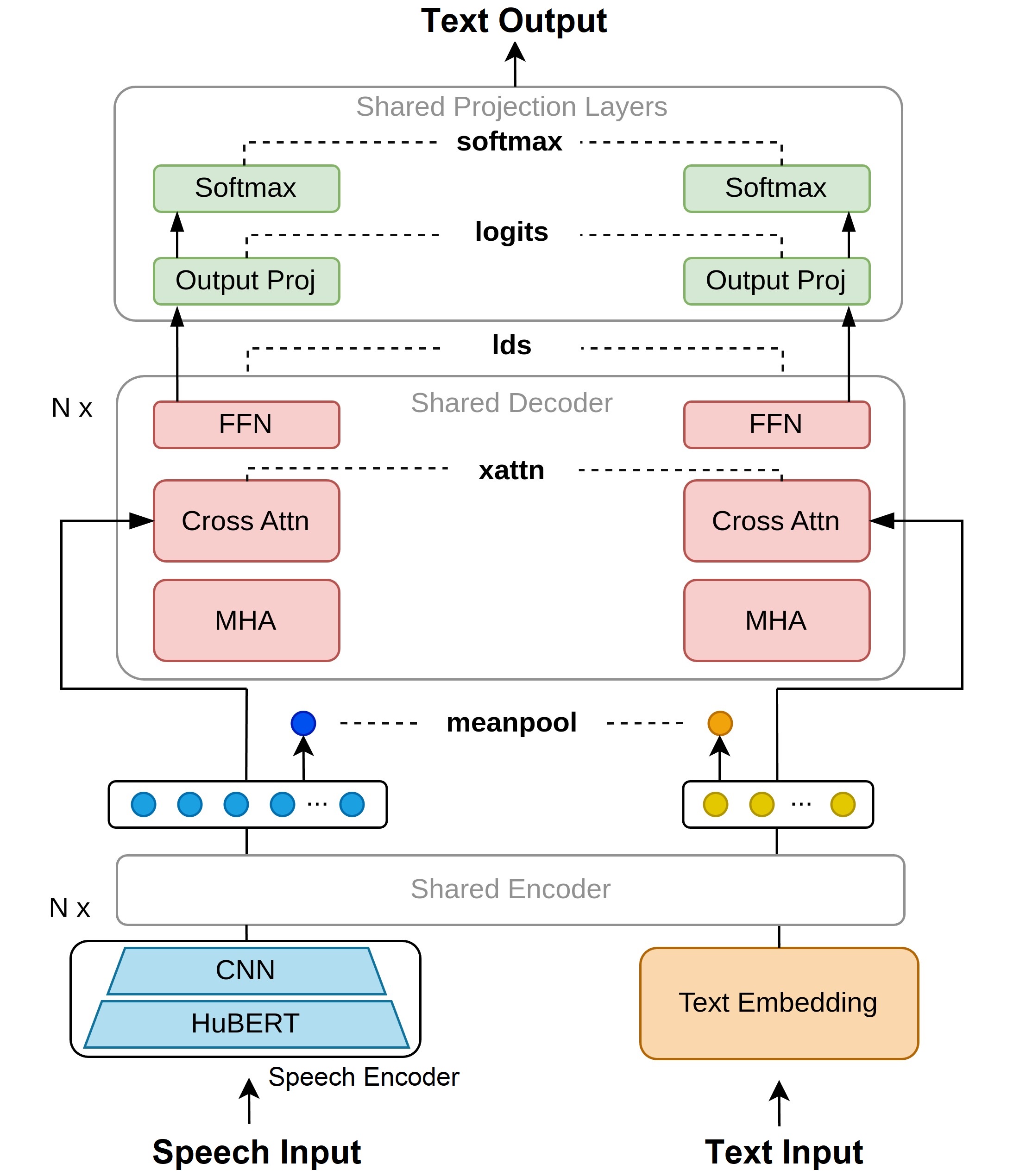}
  \caption{The MTL framework and consistency training of ST.}%
  \label{fig:consistency_where}
\end{figure}

\section{Introduction}

Speech Translation (ST) aims to translate input speech into target text in a different language. This comprises of two sub-tasks -- transcribing source speech into source text and translating source text into target text. The traditional approach involves a cascade of two models, each responsible for each sub-task \cite{bentivogli2021cascade}. An alternative is to directly translate the source speech into target text in a single model \cite{berard2016listen,berard2018end,wu2020self,xu2021stacked,barrault2023seamless, radford2023robust}. This end-to-end approach mitigates inherent problems with using two models sequentially: error propagation, increased latency and model size.

However, the end-to-end approach is limited by the scarcity of paired speech-to-text translation data. Recently, Multi-Task Learning (MTL) with Machine Translation (MT) tasks has been shown to be effective in improving the performance of an end-to-end ST model~\cite{anastasopoulos2018tied,tang2021improving,ye2021end}. Notably,~\citet{fang2022stemm,zhang2023simple,han-etal-2023-modality} train the decoder to generate consistent output for both speech and text inputs by mixing their embeddings in the latent space. \citet{han2021learning, tsiamas2023efficient} project the speech and text embeddings into a fixed-size sequence of vectors, and~\citet{ye2022cross,ouyang-etal-2023-waco,cheng2023m} align the embeddings via contrastive learning. Despite the differences in their detailed approach, they all show that we can improve the model performance by providing the model with added mechanisms to tie the distribution of the two modalities close to each other.

While it is clear that adding the auxiliary task with different modality %
is helpful for the model performance, the extra mechanisms, such as mix-ups, contrastive learning, and fixed-size vectors, may hinder our understanding on the interplay between different loss terms in MTL, and may also introduce additional inductive biases. 
For efficient training in the MTL settings while avoiding additional mechanisms, consistency training techniques, namely consistency regularization~\cite{zhang2019consistency, xie2020unsupervised} and R-drop~\cite{wu2021r}, have been proposed. 
\citet{fang-feng-2023-understanding} utilizes consistency regularization to mitigate exposure bias between ST and MT. %
\citet{gao2024empirical} explores applications of consistency regularization and R-drop for ST/MT MTL, and discovers that the former is helpful for zero-shot settings while the latter is more helpful for the regular setting. 
\citet{lee2023consistency} applies R-drop to the MT task to prevent catastrophic forgetting when fine-tuning pretrained MT models for the ST task. 
These papers successfully apply consistency training and highlight the factors that contribute to the model improvement. However, it still remains unclear as to how the individual components of the MTL affect the model training.

In this paper, we seek to explore the effects of ensuring consistency between latent representations at multiple stages of the model's forward pass and across different modalities in an MTL setting with MT task. By doing so, we shed light on the inter-dependencies of the individual loss terms in MTL that were otherwise under-studied in the existing literature, and reinforce cross-modal knowledge transfer therein. In short, our contributions are as follows:

\begin{itemize}
  \item We conduct extensive studies on various methods of consistency training and empirically verify that applying Kullback-Leibler (KL) divergence loss at the final softmax output is the most effective, regardless of the chosen consistency measure
  \item We draw similarities between different consistency measures and hypothesize that they can be combined in a unified formalism of \emph{total regularization}
  \item We find the contour with optimal total regularization (the regularization horizon) and show that the model achieves near state-of-the-art (SOTA) performance within the regularization horizon%
\end{itemize}

Unlike~\citet{gao2024empirical,lee2023consistency}, we omit the exploration of consistency within the MT task as we considered optimizing the MT task as being irrelevant to maximizing ST performance. Our design choices are specifically dedicated to understanding the effects of and the interplay between different loss terms for the purposes of maximizing the model performance on the ST task. 

\section{Method}

\subsection{Consistency Training}

We use the term consistency for two distinct purposes. Firstly, we consider the consistency between two distinct forward passes of the same speech input. Due to the dropout module, every forward pass induces different representations. Secondly, we consider the consistency between the representations of speech and text inputs along the model's forward pass. %
To avoid confusion, we refer to the former consistency as R-drop and the latter as consistency regularization.

The training data for MTL ($\mathcal D$) comprises of speech ($\mathbf{x_s}$), transcript ($\mathbf{x_t}$), and translation ($\mathbf{y}$), formally $\mathcal{D} = \{ ( \mathbf{x_s}, \mathbf{x_t}, \mathbf{y} ) \}$.
In this paper, we %
explore how consistency can be enforced in both R-drop and consistency regularization, 
by introducing the %
distance between the two forward passes %
as the auxiliary loss function.
Specifically, we define the loss function of our training objective as $\mathcal{L} = \mathcal{L}_{\textrm{ce}} + \mathcal{L}_{\textrm{con}}$, where %
$\mathcal{L}_{\textrm{ce}}$ is the cross entropy loss for ST and MT: 
\begin{align}
  \mathcal{L}_{\textrm{ce}} &= -  \alpha_s \mathbf{y} \cdot \log P( \mathbf{y} | \mathbf{x_s} ) - \alpha_t \mathbf{y} \cdot \log P( \mathbf{y} |\mathbf{x_t}).
  \label{eq:loss_tot}
\end{align}
and $\mathcal{L}_{\textrm{con}}$ is the consistency loss, which is the addition of consistency regularization loss, $\mathcal{L}^{ij}_{\textrm{cr}}$, and R-drop loss $\mathcal{L}^{ij}_{\textrm{rdrop}}$, each defined as:
\begin{align}
  \mathcal{L}_{\textrm{cr}}^{ij} &= \alpha_{\crr} D_{j}( f^i_s(\mathbf{x_s}) , f^i_t(\mathbf{x_t}) ) , \label{eq:loss-cr}\\
  \mathcal{L}_{\textrm{rdrop}}^{ij} &= \alpha_{\drr} D_{j}( f^i_s(\mathbf{x_s}) , f^i_s(\mathbf{x_s}) ).
  \label{eq:loss-rd}
\end{align}
$\alpha$'s are the loss ratio hyperparameters that we fix \(\alpha_s = 1\).
There are two additional design choices in this scheme -- which embeddings do we compare (\(f^i\)) and how do we define the distance (\(D_j\)). 
$f^i_{s/t}$ indicates speech/text embeddings at the $i$-th layer. 
We consider the encoder output (enc), cross attention (x-attn), last decoder state (lds), logits, and the softmax output as candidates for comparison,  %
depicted in Fig.~\ref{fig:consistency_where}. 

We use three different metrics to measure the distance between the embeddings.
They are the mean-square-error (MSE)
\begin{align}
  D_{\textrm {MSE}}( \mathbf{x}, \mathbf{y} ) &= \frac{1}{d} ||\mathbf{x} - \mathbf{y}||^2,
  \label{eq:mse}
\end{align}
where $d$ is the dimension of $\mathbf{x}$%
, the cosine similarity (COS)
\begin{align}
  D_{\textrm{COS}}( \mathbf{x}, \mathbf{y} ) &= 1 - \frac{\mathbf{x} \cdot \mathbf{y}}{||\mathbf{x}|| \, ||\mathbf{y}||},
  \label{eq:cos}
\end{align}
and the Kullback–Leibler (KL) divergence
\begin{align}
  D_{\textrm{KL}}( \mathbf{x}, \mathbf{y} ) &= \frac{1}{2}{ \left( \mathbf{x} \cdot \log \frac{ \mathbf{x} }{ \mathbf{y} } + \mathbf{y} \cdot \log \frac{ \mathbf{y} }{ \mathbf{x} } \right)}.
  \label{eq:kl}
\end{align}
The division and log operation of vectors are done componentwise.
Note that the KL divergence loss applied to the softmax output is in effect equal to the online knowledge distillation setting introduced in \citet{tang2021improving}.

\subsection{Experimental Setup}

\textbf{Model} \,For our speech encoder, we use HuBERT base model, pretrained\footnote{\url{https://github.com/facebookresearch/fairseq/blob/main/examples/hubert/README.md}} on 960 hours of Librispeech \cite{panayotov2015librispeech}. The HuBERT model is followed by 2 layers of convolutional subsamplers each with kernel size 5, stride 2, padding 2, resulting in 512 hidden dimensions.
The subsequent ST encoder and ST decoder are shared with the MT task and are initialized by pretraining on the MT dataset. The MT pretrained model is a 6-by-6 encoder-decoder model with 8 attention heads, 512 hidden dimensions, and 2048 feed-forward hidden dimensions.
Fig.~\ref{fig:consistency_where} illustrates this process. The entire model amounts to around 155M parameters.

\textbf{Data Preprocessing} \,For audio input, we use the raw 16kHz waveform. For text input, we use the tokenizer trained on both the transcription and translation of our ST dataset with a vocabulary size of 10k using unigram in SentencePiece \cite{kudo2018sentencepiece}.

\textbf{Dataset} \,We use the MuST-C \cite{di2019must} dataset, and our main investigation on consistency training is conducted on German (De). We then apply our best method to Spanish (Es), French (Fr) and Italian (It)\footnote{We use v2.0 for De and v1.0 for Es, Fr, and It}.
The transcription and translation pairs in MuST-C dataset are used as our MT dataset. We also use WMT \cite{buck2016findings} for languages (De/Es/Fr) and OPUS100 \cite{zhang2020improving} for It as our external MT dataset during pre-training of the shared encoder/decoder components.

\textbf{Training} \,We train our models using the \texttt{fairseq}~\cite{ott2019fairseq} framework. We use Adam~\cite{2015-kingma} as our optimizer for training. The learning rate is scheduled using an inverse square root scheduler with maximum learning rate of $1 \times 10^{-4}$ and $4000$ warm-up steps. FP16 training is used as provided in fairseq, without gradient clipping. We train our models on 2 Nvidia V100 GPUs with gradient accumulation of 4 steps, leading to an effective batch size of up to 16 million audio tokens. R-drop requires additional GPU space, as such, we half the batch size for each gradient accumulation and use 4 V100 GPUs to preserve the effective batch size.

\textbf{Evaluation} \,Our models are validated using case-sensitive detokenized BLEU scores \cite{papineni2002bleu} on MuST-C \texttt{dev} set using sacreBLEU v1.5.1 \cite{post2018call}. We stop training when the validation BLEU score does not improve for 10 checkpoints, and the averaged last 10 checkpoints is selected as our final model. We report our scores using the same BLEU score metric on \texttt{tst-COMMON} of MuST-C. We use paired bootstrap resampling \cite{koehn2004statistical} to compute the statistical significance of our results and report them together with the BLEU scores in Table~\ref{tab:full}. All reported scores achieve statistical significance ($p < 0.05$) over the baseline with paired bootstrap resampling unless noted with ($^\dagger$).

\setlength{\tabcolsep}{2.5pt} %
\begin{table}[t]
  \centering
  
  \resizebox{\columnwidth}{!}{%
  \begin{tabular}{ l l l c c}
    \toprule
    & \textbf{Experiment} & \textbf{Loss} & \textbf{Base} & \textbf{Expand} \\
    \midrule
    \midrule
    & baseline            & $\mathcal{L}_{\textrm{ce}}$ & 25.77 & 28.20 \\
    \midrule
    \multirow{9.6}{*}{\makecell[c]{con.\\reg.}} & enc-MSE        & $\mathcal{L}_{\textrm{ce}}+ \mathcal{L}_\crr^{\textrm{encMSE}}$  & 26.01$^\dagger$ & 28.35$^\dagger$ \\ %
    &enc-COS        & $\mathcal{L}_{\textrm{ce}}+\mathcal{L}_\crr^{\textrm{encCOS}}$  & 25.93$^\dagger$ & 28.32$^\dagger$ \\ %
    \cmidrule{2-5}
    &xattn-MSE        & $\mathcal{L}_{\textrm{ce}}+ \mathcal{L}_\crr^{\textrm{xattnMSE}}$  & 25.93$^\dagger$ & 28.45$^\dagger$ \\
    &xattn-COS        & $\mathcal{L}_{\textrm{ce}}+ \mathcal{L}_\crr^{\textrm{xattnCOS}}$  & 26.11 & 28.72 \\

    &lds-MSE        & $\mathcal{L}_{\textrm{ce}}+ \mathcal{L}_\crr^{\textrm{ldsMSE}}$  & 26.39 & 28.89 \\
    &lds-COS        & $\mathcal{L}_{\textrm{ce}}+ \mathcal{L}_\crr^{\textrm{ldsCOS}}$  & 26.33 & 28.67 \\
    \cmidrule{2-5}
    &logits-MSE        & $\mathcal{L}_{\textrm{ce}}+ \mathcal{L}_\crr^{\textrm{logitsMSE}}$  & 26.78 & 28.88 \\
    \cmidrule{2-5}
    &softmax-MSE       & $\mathcal{L}_{\textrm{ce}}+ \mathcal{L}_\crr^{\textrm {softmaxMSE}}$  & \textbf{27.45} & 29.34 \\
    &softmax-KL        & $\mathcal{L}_{\textrm{ce}}+ \mathcal{L}_\crr^{\textrm{softmaxKL}}$  & 27.32 & \textbf{29.59} \\
    \midrule
    \multirow{6.8}{*}{\makecell[c]{R-drop}} &lds-MSE        & $\mathcal{L}_{\textrm{ce}}+ \mathcal{L}_\drr^{\textrm{ldsMSE}}$  & 26.96 & 28.84 \\
    &lds-COS        & $\mathcal{L}_{\textrm{ce}}+ \mathcal{L}_\drr^{\textrm{ldsCOS}}$  & 26.44 & 28.30$^\dagger$ \\
    \cmidrule{2-5}
    &logits-MSE        & $\mathcal{L}_{\textrm{ce}}+ \mathcal{L}_\drr^{\textrm{logitsMSE}}$  & 27.12 & 28.79 \\
    \cmidrule{2-5}
    &softmax-MSE       & $\mathcal{L}_{\textrm{ce}}+ \mathcal{L}_\drr^{\textrm {softmaxMSE}}$  & 27.49 & 29.45 \\
    &softmax-KL        & $\mathcal{L}_{\textrm{ce}}+ \mathcal{L}_\drr^{\textrm{softmaxKL}}$  & \textbf{28.40} & \textbf{29.66} \\
    \bottomrule
  \end{tabular}
  }
  \caption{BLEU scores with various consistency regularization and R-drop on MuST-C en-de dataset.}
  \label{tab:const_experiments}
\end{table}
\setlength{\tabcolsep}{6.pt} %

\section{Results}
\label{sec:results}

\setlength{\tabcolsep}{2.5pt} %
\begin{table}[t]
  \centering
  \resizebox{\columnwidth}{!}{%
  \begin{tabular}{l l c c}
    \toprule
    \multicolumn{1}{c}{\textbf{Experiment}} & \multicolumn{1}{c}{\textbf{Loss}} & \multicolumn{1}{c}{\textbf{Base}} & \multicolumn{1}{c}{\textbf{Expand}} \\
    \midrule
    \midrule
    baseline            & $\mathcal{L}_{\textrm{ce}} + \mathcal{L}_\drr^{\textrm{softmaxKL}}$ & 28.40 & 29.66 \\
    \midrule
    enc-MSE        & $\mathcal{L}(\textrm{baseline})+ \mathcal{L}_\crr^{\textrm{encMSE}}$  & 28.07 & 29.38 \\
    enc-COS        &  $\mathcal{L}(\textrm{baseline})+\mathcal{L}_\crr^{\textrm{encCOS}}$  & 28.15 & 29.41 \\
    \midrule
    xattn-MSE        &  $\mathcal{L}(\textrm{baseline})+ \mathcal{L}_\crr^{\textrm{xattnMSE}}$  & 28.14 & 29.64 \\
    xattn-COS        &  $\mathcal{L}(\textrm{baseline})+ \mathcal{L}_\crr^{\textrm{xattnCOS}}$  & 28.08 & 29.43 \\
    \midrule
    lds-MSE        &  $\mathcal{L}(\textrm{baseline})+ \mathcal{L}_\crr^{\textrm{ldsMSE}}$  & 28.19 & 29.62 \\
    lds-COS        &  $\mathcal{L}(\textrm{baseline})+ \mathcal{L}_\crr^{\textrm{ldsCOS}}$  & \textbf{28.42} & 29.50 \\
    \midrule
    logits-MSE        &  $\mathcal{L}(\textrm{baseline})+ \mathcal{L}_\crr^{\textrm{logitsMSE}}$  & 28.39 & 29.43 \\
    \midrule
    softmax-MSE       &  $\mathcal{L}(\textrm{baseline})+ \mathcal{L}_\crr^{\textrm {softmaxMSE}}$  & 28.14 & 29.61 \\
    softmax-KL        &   $\mathcal{L}(\textrm{baseline})+\mathcal{L}_\crr^{\textrm{softmaxKL}}$  & 28.37 & \textbf{29.73} \\
    \bottomrule
  \end{tabular}
  }
  \caption{BLEU scores with both consistency regularization and R-drop on MuST-C en-de dataset. %
  }
  \label{tab:cr+rdrop}
\end{table}
\setlength{\tabcolsep}{6.pt} %

\subsection{Consistency Regularization and R-Drop}

We first investigate the effects of consistency regularization when varying the compared embeddings and the distance metrics (the $i$ and $j$'s of~\eqref{eq:loss-cr}).
The BLEU scores for the en-de ST task are shown in Table~\ref{tab:const_experiments}.
Note that the expand setting utilizes the external MT dataset during pre-training.

The results suggest that introducing consistency regularization indeed enhances the performance of the ST, regardless of the selected embedding or metric.
Moreover, enforcing consistency between embeddings that are closer to the final output layer results in better performance for both base and expanded settings. 
This is consistent with \citet{pham2019improving}, where they note that enforcing consistency on the embeddings that are nearest to the final output layer offers most freedom for the model to optimize on the most ideal internal representations.

The difference between $D_{\textrm{MSE}}$ and $D_{\textrm{COS}}$ did not show any clear trend. %
Since MSE depends on both the norm of the two vectors and the angle between, while COS is solely a function of the angle, this suggests that the consistency is imposed mainly through \emph{aligning} the embedding vectors, and restricting the norm does not provide additional regularization.

For the softmax output of the network, we apply MSE in addition to the KL divergence, 
motivated by the fact that the softmax-KL performed much better than the logits-MSE.
Assuming the logit outputs of speech and text are $\mathbf{x}$ and $\mathbf{y}$, then $\mathcal{L}_\crr^{\textrm{logitsMSE}} \propto D_{\textrm{MSE}}( {\mathbf{x}}, {\mathbf{y}} ) \propto ( \mathbf{x} - \mathbf{y} ) \cdot (\mathbf{x} - \mathbf{y})$. %
The softmax output is the normalized exponential of ${\mathbf{x}}$, ${\mathbf{y}}$, and thus $\mathcal{L}_\crr^{\textrm{softmaxKL}} \propto 
 D_{\textrm{KL}}( {\textrm{softmax}(\mathbf{x}}), \textrm{softmax}({\mathbf{y}}) )$ roughly scales as $( \mathbf{x} - \mathbf{y} ) \cdot (e^\mathbf{x} - e^\mathbf{y})$. 
The exponentiation amplifies the distance between the two vectors and thus has the effect of tying the two embeddings more strongly. 
We tried achieving even stronger alignment between the two vectors with $\mathcal{L}_\crr^{\textrm{softmaxMSE}} \propto 
 D_{\textrm{MSE}}( {\textrm{softmax}(\mathbf{x}}), \textrm{softmax}({\mathbf{y}}) )$, that roughly scales as $(e^\mathbf{x} - e^\mathbf{y}) \cdot (e^\mathbf{x} - e^\mathbf{y})$. 

However, the difference between softmax-MSE and softmax-KL was less prominent than that of softmax-KL and logits-MSE. %
We choose the more conventional softmax-KL as our best model and consider it as the baseline in the following sections.

We perform similar experiments in R-drop with various embeddings and distance metrics, 
observing similar trends in Table~\ref{tab:const_experiments}. %
R-drop enhances the performance for all embeddings and distance metrics, and optimizing on the embeddings that more directly impact the final output leads to better performance. 

The R-drop best results were better than consistency regularization. 
The gap is quite significant in the base setting, which indicates that with such limited data imposing consistency directly to the target ST task is a more efficient use of the resources. 
In contrast, when the large external MT dataset is available, enforcing the ST-MT consistency substantially narrows the gap and the two methods show little difference in performance.

\begin{table}[t]
  \centering
    \resizebox{\columnwidth}{!}{%
  \begin{tabular}{ l r r}
    \toprule
    \multicolumn{1}{c}{\textbf{Experiment}} & \multicolumn{1}{c}{\textbf{Base}} & \multicolumn{1}{c}{\textbf{Expand}}\\
    \midrule
    \midrule
    baseline 1 ($\mathcal{L}_{\textrm{ce}} + \mathcal{L}_\crr^{\textrm{softmaxKL}}$)           &  27.32  & 29.59\\ %
    \midrule
    $~~+ (\alpha_t = 0.5)$   & 27.41   & 29.34\\ %
    $~~+ (\alpha_t = 0.1)$   & \textbf{27.68} & \textbf{29.68}   \\ %
    $~~+ (\alpha_t = 0.0)$   & 27.44 & 29.64   \\ %
    \midrule
    baseline 2 ($\mathcal{L}_{\textrm{ce}} + \mathcal{L}_\crr^{\textrm{softmaxKL}}+\mathcal{L}_\drr^{\textrm{softmaxKL}}$)           &  28.37  & 29.73\\ %
    \midrule
    $~~+ (\alpha_t = 0.5)$   & \textbf{28.56}   & 29.72\\ %
    $~~+ (\alpha_t = 0.1)$   & 28.10 & 29.37   \\ %
    $~~+ (\alpha_t = 0.0)$   & 28.18 & 29.30   \\ %
    \bottomrule
  \end{tabular}
  }
  \caption{BLEU score with various combinations of regularization to the ST model. %
  }
  \label{tab:reinforce_kl}
\end{table}
\setlength{\tabcolsep}{6.pt} %

\setlength{\tabcolsep}{3.5pt}
\begin{table*}[t]
  \centering
  \resizebox{0.9\textwidth}{!}{%
  \begin{tabular}{ l c c c c | c || c c c c | c }
    \toprule
    \multirow{2}{*}{\makecell[c]{\textbf{Model}}} & \multicolumn{5}{c}{\textbf{Base}} & \multicolumn{5}{c}{\textbf{Expand}} \\
    \cmidrule{2-6} \cmidrule{7-11}
    & \textbf{en-de} & \textbf{en-es} & \textbf{en-fr} & \textbf{en-it} & \textbf{avg} 
    & \textbf{en-de} & \textbf{en-es} & \textbf{en-fr} & \textbf{en-it} & \textbf{avg} \\
    \midrule
    XSTNet \cite{ye2021end} & 25.5 & 29.6 & 36.0 & 25.5 & 29.2 & 27.8 & 30.8 & 38.0 & 26.4 & 30.8 \\
    ConST \cite{ye2022cross} & 25.7 & 30.4 & 36.8 & 26.3 & 29.8 & 28.3 & 32.0 & 38.3 & 27.2 & 31.5 \\
    FCCL \cite{zhang2023improving} & 25.9 & 30.7 & 36.8 & 26.4 & 30.0 & 29.0 & 31.9 & 38.3 & 27.3 & 31.6 \\
    $\text{M}^3$ST \cite{cheng2023m} & 26.4 & 31.0 & 37.2 & 26.6 & 30.3 & 29.3 & 32.4 & 38.5 & 27.5 & 31.9 \\
    CRESS \cite{fang-feng-2023-understanding} & 27.2 & 31.9 & 37.8 & 27.3 & 31.1 & 29.4 & 33.2 & \textbf{40.1} & 27.6 & 32.6 \\
    Consistency is Key \cite{lee2023consistency} & – & – & – & – & – & 29.3 & 32.3 & 39.5 & – & – \\
    SimRegCR \cite{gao2024empirical} & 27.9 & \textbf{32.1} & 39.0 & \textbf{27.7} & \textbf{31.7} & 29.2 & 33.0 & 40.0 & \textbf{28.2} & 32.6 \\
    Hard Multi-task (190M) \cite{yan2024cross} & – & – & – & – & – & \textbf{30.1} & 33.2 & 39.2 & – & – \\
    \midrule
    Ours &&&&& &&&&& \\
    baseline 2 + ($\alpha_{\crr}=5$) + d/o=0.05 & 28.2 & 31.5  & 37.5 & 26.8 & 31.0 & 29.9 & \textbf{33.4} & 39.7 & 27.4 & 32.6 \\
    baseline 2 + ($\alpha_{\drr}=8$) + d/o=0.05 & 28.0 & 31.6 & 38.4 & 27.1 & 31.3 & 29.9 & \textbf{33.4} & 39.9 & 27.5 & \textbf{32.7} \\
    \bottomrule
  \end{tabular}
  }
  \caption{BLEU scores across 4 languages in MuST-C dataset, compared with other baseline models in the literature.}
  \label{tab:full}
  \vspace{-7pt}
\end{table*}
\setlength{\tabcolsep}{6pt}

\subsection{Compounding Consistency Losses}

We now turn to the problem of employing both regularization schemes simultaneously. 
To observe as many data points as possible, we choose $\mathcal{L}_{\textrm{ce}} + \mathcal{L}_\drr^{\textrm{softmaxKL}}$ as the baseline and add each consistency regularization loss shown in Table~\ref{tab:const_experiments} in tandem.
These results are presented in Table~\ref{tab:cr+rdrop}.

We notice that the BLEU scores of the experiments are surprisingly stable compared to Table~\ref{tab:const_experiments}.
This suggests that the two regularizations does not simply add up but rather saturates. %
We conjecture the reason for this behavior stems from the inherent similarity between consistency regularization and R-drop.
As seen in \eqref{eq:loss-cr} and \eqref{eq:loss-rd}, the only difference is comparing speech-text embeddings and speech-speech embeddings. 
Having loss terms with the same form but with different modality is imposing the same \emph{type} of regularization. 
In a similar concept called the ``$m$-time R-drop''~\cite{wu2021r}, which ensures consistency between $m$ sub-models (the conventional R-drop is a 2-time R-drop), increasing $m$ does not improve the performance but rather quickly saturates.
This is analogous to the observed saturation as compounding the two consistencies is essentially a 3-time R-drop where one sub-model is of text modality. 

\subsection{Coefficient of MT loss}

We ablate $\alpha_t$ using two different baselines in Table~\ref{tab:reinforce_kl} to observe the effects. 
While tuning $\alpha_t$ can be construed as a hyperparameter search, our experiments show that decreasing $\alpha_t$ decreases the relative strength of the cross-entropy loss, which effectively increases regularization. 
We observe that by decreasing $\alpha_t$ from 1.0 to 0.0, the BLEU score mostly increases initially but eventually decreases, which is a typical behavior of increasing regularization.
We elaborate on the regularization aspect of tuning $\alpha_t$ in the Discussion section. 

For the baseline 1, the peak performance occurred at $\alpha_t = 0.1$.
For the baseline 2, the BLEU score peaked at $\alpha_t = 0.5$ for base setting and decreased monotonically for the expanded setting (there is, of course a possibility that the peak of the expanded setting has occurred between $0.5 <\alpha_t< 1.0$).
This is consistent with our argument since with the R-drop loss, baseline 2 already has larger regularization than the baseline 1. 
Therefore when decreasing $\alpha_t$ the total regularization is greater in the baseline 2, and would have a peak at a larger value of $\alpha_t$.

\section{Discussion}

\begin{figure*}[t]
  \centering
  \includegraphics[width=0.32\textwidth]{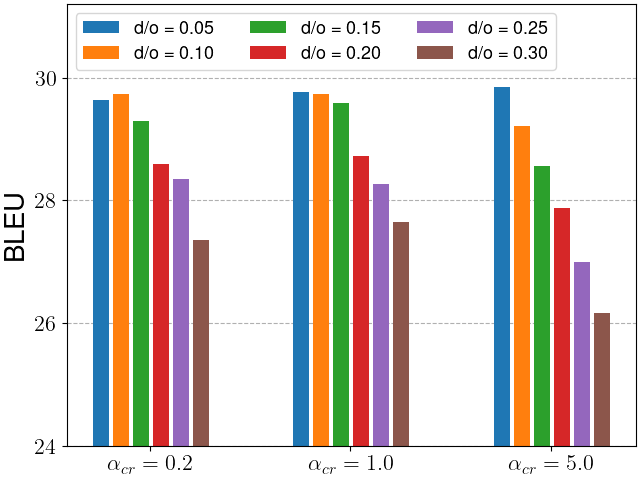}
  \includegraphics[width=0.32\textwidth]{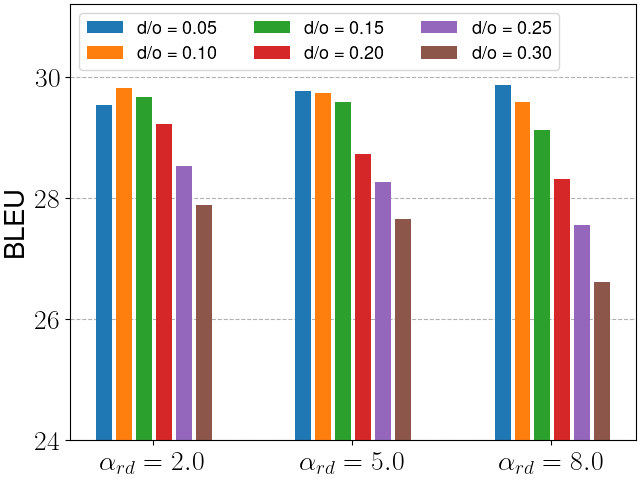}
  \includegraphics[width=0.32\textwidth]{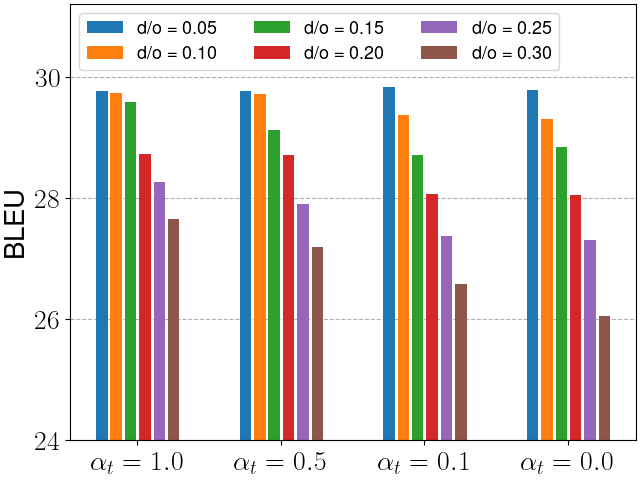}
  \caption{
  The effect of dropout rate on the ST BLEU score, when varying the loss rate coefficients. The form of the loss function is that of the baseline 2 in Table~\ref{tab:reinforce_kl}. The default values are $\alpha_{\crr} = 1.0$, $\alpha_{\drr} = 5.0$, and $\alpha_{t} = 1.0$. One can again verify that decreasing $\alpha_t$ has the same effect as increasing $\alpha_{cr/rd}$, which is increasing regularization. 
  }
  \label{fig:regularization_horizon_prev}
\end{figure*}

\subsection{Consistency in MTL as Means of Regularization}

Here, we give an alternative explanation of how decreasing $\alpha_t$ serves as a regularization in the context of the analogy between consistency regularization and R-drop.

As MTL benefits from the transfer of knowledge between tasks, it also improves the model's generalizability \cite{caruana1997multitask}. 
In the MTL setting with shared outputs, we can consider the text input for the MT task as a corrupted version of the speech input.
Consistency regularization can be thought of as encouraging model consistency between a normal input and a corrupted input -- similar in spirit to the Cutoff approach~\cite{shen2020simple}. 
The consistency regularization objective function thereby amplifies the cost of overfitting and enhances the model's generalizability.
Within this context, R-drop and consistency regularization resemble each other; both regularize the inconsistency between the outputs of a corrupted forward pass (either the model is corrupted or the input is corrupted). This is also mentioned in~\citet{wu2021r} as they describe the similarity between R-drop and the Cutoff.

With the consistency regularization in place, $\alpha_t$ becomes the determinant of how corrupted the output of MT forward pass is. A lower $\alpha_t$ value yields less optimal outputs, which can be considered as increased perturbations to the input, effectively increasing the regularization. 
The role of $\alpha_t$ in consistency regularization is analogous to the role of the dropout rate in R-drop. Increasing the dropout rate in R-drop achieves greater perturbation in the output, yielding a stronger regularization.

\subsection{The Regularization Horizon}

As we have established $\alpha_t$ as a knob for regularization, we can now collect all sources of regularization in our scheme -- consistency regularization, R-drop, and $\alpha_t$, and define the concept of \emph{total regularization}.
Including the dropout rate (d/o), there are four parameters which controls the total regularization ($R$): $R = f(\alpha_{\crr}, \alpha_{\drr}, \alpha_t, \textrm{d/o})$.

Determining the analytic form of $f$ is a formidable task.
However, from our experiments, we can infer several core properties of the function.
First, $f$ monotonically increases with $\alpha_{\crr}$, $\alpha_{\drr}$, and d/o, and monotonically decreases with $\alpha_t$.
Second, while $\alpha_t$ and d/o have little correlation with others (that is, the composite effect with other parameters easily adds up for those parameters), the effect of having both $\alpha_{\crr}$ and $\alpha_{\drr}$ saturates as seen in Table~\ref{tab:cr+rdrop}. 

In general, regularization has some optimal value. 
Some regularization benefits the model by preventing overfitting, but excessive regularization hurts the model's performance on its original task. 
Therefore, the performance will increase to a peak value and then decrease as one increases regularization.
With the total regularization as a function of four parameters, the optimal strength of regularization would be represented as a three-dimensional surface in the four-dimensional parameter space. 
We refer to this optimal surface as the \emph{regularization horizon}, beyond which the model performance begins to collapse rapidly.

In order to approximate the relationship between the regularization forces and the resulting total regularization, we %
first measure the BLEU scores of the baseline 2 in Table~\ref{tab:reinforce_kl} while tuning the dropout rate and $\alpha_{\crr/\drr/t}$. 
From the default value $\alpha_{\crr} = 1.0$, $\alpha_{\drr} = 5.0$, and $\alpha_{t} = 1.0$, we separately tune $\alpha_{\crr} \in \{0.2, 1.0, 5.0\}$, $\alpha_{\drr} \in \{2.0, 5.0, 8.0\}$ and $\alpha_t \in \{1.0, 0.5, 0.1, 0.0\}$. 
For each set of $\alpha$'s, we plot the BLEU scores as a function of the dropout rate, ranging from 0.05 to 0.30. 
The result of this experiment is shown in Fig.~\ref{fig:regularization_horizon_prev}.
The plots consistently show that the BLEU score approaches peak performance at the lower end of the dropout rate and then begins to monotonically decrease after the peak as the dropout rate increases.
The monotonically decreasing plots for large $\alpha_{\crr}$, $\alpha_{\drr}$ and small $\alpha_t$ indicates the data with $0.05$ dropout rate already has large enough regularization and passed the peak performance. 
Additionally, the similar behavior of $\alpha_t$ to the other two panels is another evidence that decreasing $\alpha_t$ has the same effect as increasing $\alpha_{\crr}$ or $\alpha_{\drr}$.

Now, we assume a linear function for $f$, that is:
\begin{align}
    R = \beta_{\crr} \alpha_{\crr} + \beta_{\drr}  \alpha_{\drr} + \beta_{t}  \alpha_t + \beta_{\text{do}} ( \text{d/o}) + \beta_f .
\end{align}
While this is evidently an oversimplification, it serves as a good starting point for demonstrating the total regularization and regularization horizon.
We can also consider this as a Taylor expansion of $R$ in the four parameters, $(\alpha_{\crr}, \alpha_{\drr}, \alpha_t, \textrm{d/o})$, and approximating it up to linear terms. 
We make another approximation that the BLEU score decreases linearly with $R$ in the regime of excessive regularization:
\begin{align}
    BLEU = \beta_R  R + \beta_{B}.
    \label{eq:reg2}
\end{align}
This is a reasonable approximation within our parameter range of interest, as can be seen in $\alpha_\crr = 5.0$ and $\alpha_t = 0.1$ plots in Fig.~\ref{fig:regularization_horizon_prev}.

Combining the two linear approximations, we regress the BLEU score on the $\alpha$'s and d/o\footnote{\label{reg_coeff_val}The resulting regression coefficients are: $\beta_{\crr} = 0.245$, $\beta_{\drr}=0.159$, $\beta_{t}=-0.814$, $\beta_{\text{do}}=13.8$, $\beta_f = 0.814$,  $\beta_B = 32.6$.}.
As \eqref{eq:reg2} is only valid in the over-regularization regime, we select the points \emph{after} the peak in Fig.~\ref{fig:regularization_horizon_prev} for the regression. 
The magnitude of $\beta_R$ merely sets the scale of $R$, and we fix this as $\beta_R = -1$.
We also fix the ambiguity between $\beta_f$ and $\beta_B$ by defining $f(0,0,1,0) = 0$.

\begin{figure}[t]
  \centering
  \includegraphics[height=4.8cm]{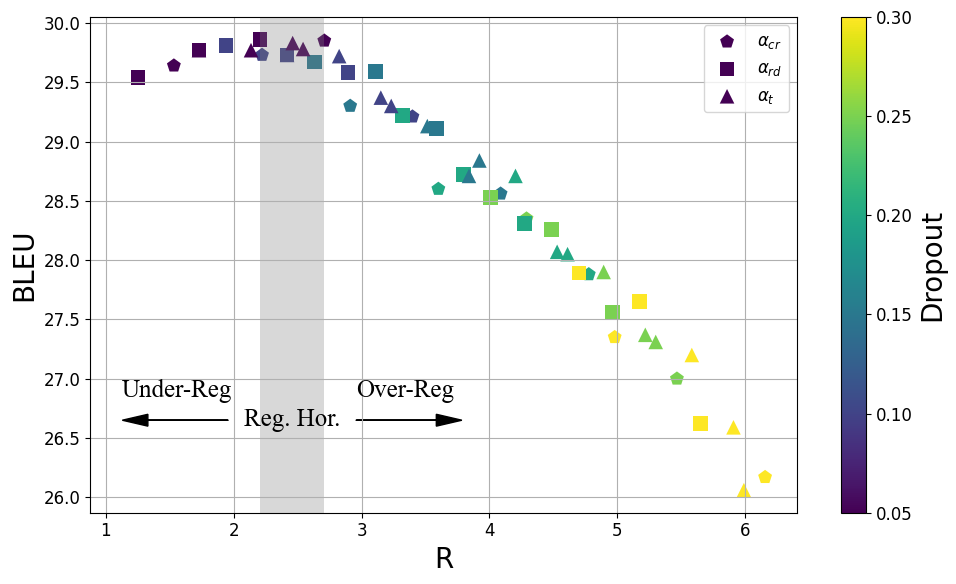}
  \caption{
  We plot the BLEU score against the total regularization, $R$, obtained from the regression result. Different symbols corresponds to the tuned $\alpha$ (the panels in Fig.~\ref{fig:regularization_horizon_prev}). %
  The color scheme indicates the dropout rate of each data. 
  The regularization horizon and under/over-regularized regions are indicated, respectively.
  }
  \label{fig:regularization_horizon}
\end{figure}

We assign $R$ values to each experiment from the regression coefficients and plot the BLEU score as a function of $R$ for all data in Fig.~\ref{fig:regularization_horizon}.
While the three graphs with distinct symbols tune different $\alpha$'s, one can observe that they collapse to a single curve, even for the points not included in the regression. 
This is strong evidence that total regularization is a valid variable that controls the overall performance. 
The regularization horizon is placed at the peak region of the figure (shaded in gray), where the regularization is optimal.
We define the region with less than optimal regularization as the \emph{under-regularized} regime and more than optimal as the \emph{over-regularized} regime.
Note that only the points in the over-regularized regime were used in the regression.

\subsection{Multilingual Speech Translation}

We identify several points on the regularization horizon, which are combinations of consistency terms that maximize the model performance within the architectural framework of MTL. 
Finally, we apply our findings across 4 languages (De/Es/Fr/It) using the expanded setting and observe that they are consistent across different languages. As reported in Table~\ref{tab:full}, our model achieves competitive performance with SOTA methods.

\section{Conclusion}

In this paper, we present a systematic exploration of consistency training methods for improving ST within a MTL framework. 
We show that applying consistency constraints -- whether across modalities via consistency regularization or within modalities via R-drop -- is most effective when applied closer to the model’s output layer. 
Additionally, we identify the MT loss weight as a tunable source of regularization, functioning analogously to dropout rate in R-drop. 
By unifying these components, we introduce the concept of \emph{total regularization} and empirically define a \emph{regularization horizon}, which represents an optimal region in the hyperparameter space where model performance peaks. 
Our experiments demonstrate that operating near this horizon consistently leads to strong performance across multiple languages, achieving results competitive with state-of-the-art methods on the MuST-C benchmark. 
These findings offer new insights into how the MTL dynamics can be framed and optimized through regularization.

\bibliography{custom}

\end{document}